\documentclass[preprint, prX]{revtex4}

\usepackage{amsmath}    % need for subequations
\usepackage{graphicx}   % need for figures
\usepackage{verbatim}   % useful for program listings
\usepackage{color}      % use if color is used in text
\usepackage{subfigure}  % use for side-by-side figures
\usepackage{hyperref}   % use for hypertext links, including those to external documents and URLs
\usepackage{amssymb}    % Necessary for \Bbb{R}, etc.
\usepackage{epsfig}
\usepackage{graphics,graphicx}
\usepackage{setspace}
\usepackage{url}
\usepackage{algorithm,algorithmic}
\usepackage{amssymb}

\usepackage{multirow}
\usepackage{tabularx}

\begin{document}

\begin{abstract}
The Artificial Bee Colony (ABC) is the name of an optimization algorithm that was inspired by the intelligent behavior of a honey bee swarm. It is widely recognized as a quick, reliable, and efficient methods for solving optimization problems. This paper proposes a hybrid ABC (HABC) algorithm for graph 3-coloring, which is a well-known discrete optimization problem. The results of HABC are compared with results of the well-known graph coloring algorithms of today, i.e. the Tabucol and Hybrid Evolutionary algorithm (HEA) and results of the traditional evolutionary algorithm with SAW method (EA-SAW). Extensive experimentations has shown that the HABC matched the competitive results of the best graph coloring algorithms, and did better than the traditional heuristics EA-SAW when solving equi-partite, flat, and random generated medium-sized graphs.

\textit{To cite paper as follows: I.Jr.~Fister, I.~Fister and J.~Brest. A Hybrid Artificial Bee Colony Algorithm for Graph 3-Coloring. In { \em Swarm and Evolutionary Computation}, Lecture Notes in Computer Science, 7269, Springer Berlin / Heidelberg, 66--74 (2012).
}

\end{abstract}

\title{A Hybrid Artificial Bee Colony Algorithm\\for Graph 3-Coloring}

\author{Iztok Fister Jr.}
\altaffiliation{University of Maribor, Faculty of electrical engineering and computer science
Smetanova 17, 2000 Maribor}
\email{iztok.fister@guest.arnes.si}

\author{Iztok Fister}
\altaffiliation{University of Maribor, Faculty of electrical engineering and computer science
Smetanova 17, 2000 Maribor}
\email{iztok.fister@uni-mb.si}

\author{Janez Brest}
\altaffiliation{University of Maribor, Faculty of electrical engineering and computer science
Smetanova 17, 2000 Maribor}
\email{janez.brest@uni-mb.si}

\maketitle

\section{Introduction}
Graph coloring represents a test bed for many newly developed algorithms because of its simple definition, which states: How to color a graph $G$ with the $k$ colors, so that none of the vertices connected with an edge have the same color. The coloring $c$ is $proper$ if no two connected vertices are assigned to the same color. A graph is $k$-$colorable$ if it has a proper $k$-$coloring$. The minimum $k$ for which a graph $G$ is $k$-$colorable$ is called its $chromatic\ number\ \chi(G)$.

Many approaches for solving the graph coloring problem (GCP) have been proposed over the time~\cite{Galinier:2006,Malaguti:2009}. The most natural way to solve this problem is, however, in a greedy fashion, where the vertices of the graph are ordered into a permutation, and colored sequential. Thus, the quality of coloring depends on the permutation of the vertices. For example, the DSatur algorithm~\cite{Brelaz:1979}, one of the best traditional heuristics for graph coloring today, orders the vertices $v$ according to  $saturation\ degrees\ \rho(v)$. The saturation degree represents the number of distinctly colored vertices adjacent to the vertex $v$. Furthermore, DSatur's ordering is calculated dynamically during the coloring process.

Many heuristic methods have been developed for larger instances~\cite{Galinier:2006} because exact algorithms can only color instances of up to 100 vertices. These methods can be divided into local search methods~\cite{Aarts:1997} and hybrid algorithms~\cite{Lu:2010}. The most important representative of the former is Tabucol~\cite{Hertz:1987}, which utilizes the tabu search, as proposed by Glover~\cite{Glover:1986}. Later were combined local search methods with evolutionary algorithms and improved the results of pure Tabucol, as for example, the hybrid genetic algorithm by Fleurent and Ferland~\cite{Fleurent:1996}, and the hybrid evolutionary algorithm (HEA) by Galinier and Hao~\cite{Galinier:1999}.

Swarm intelligence is the collective behavior of a self-organized system. Birds, insects, ants, and fish use collective behavior for foraging and defending. These individuals are looking for good food sources and help each other when a lack of food has arisen. This concept was introduced into the computer's world by Kennedy and Eberhart~\cite{Kennedy:1995}. Moreover, it was successfully applied to several problem domains, for example, particle swarm optimization, which achieves good results during antenna optimization~\cite{Robinson:2004}. In addition, ant colony optimization reaches good results by solving the traveling-salesman person~\cite{Dorigo:2004}. Finally, the artificial bee colony algorithm, proposed by Karaboga and Basturk~\cite{Karaboga:2009}, exhibited excellent results when solving combinatorial optimization problems~\cite{Pan:2011,Tasgetiren:2011}.

This paper focuses on the artificial bee colony (ABC) algorithm for graph 3-coloring (3-GCP), which belongs to a class of $NP$-$complete$~\cite{Garey:1979} problems. There, the real-valued weights $w$ are assigned to the vertices $v$. These weights determine how difficult the vertex is to color. The higher the weight, the earlier the vertex should be colored. Thus, weights define the order in which the vertices should be colored. This ordering is used by the DSatur traditional algorithm for constructing 3-coloring. The ABC algorithm incorporates DSatur as a decoder. In this manner, the ABC algorithm acts as a meta-heuristic concerned for a generation of  new solutions (vector of weights), whilst the quality of the solution (its fitness) is evaluated by DSatur. This approach is not new: it was used by the evolutionary algorithm with SAW method (EA-SAW) of Eiben et al.~\cite{Eiben:1998}, and by the hybrid self-adaptive differential evolution of Fister et al.~\cite{Fister:2011}. In the former case, instead of Dsatur, a greedy heuristic was applied as a decoder. Finally, the proposed ABC algorithm was hybridized with a $random\ walk\ with\ direction\ exloitation$ (RWDE)~\cite{Rao:2009} local search heuristic. This local search heuristic was applied in place of the original sending scouts function and focuses itself on discovering new food sources in the vicinity of the current sources.

The results of the proposed hybrid artificial bee colony algorithm for graph 3-coloring (HABC) was compared with the results obtained with Tabucol, HEA, and EA-SAW for solving an extensive set of random medium-scale graphs generated by the Culberson graph generator~\cite{Culberson:2011}. A comparison between these algorithms shows that the results of the proposed HABC algorithm are comparable with results of the other algorithms used in the experiments.

The structure of this paper is as follows: In Section 2, the 3-GCP is discussed, in detail. The HABC is described in Section 3, whilst the experiments and results are presented in Section 4. The paper is concluded with a discussion about the quality of the results, and directions for further work are outlined.

\section{Graph 3-coloring}
$3$-$coloring$ of a graph $G=(V,E)$ is a mapping $c:V \rightarrow C$, where $C=\{1,2,3\}$ is a set of three colors~\cite{Bondy:2008}. Note that $V$ in the graph definition denotes a set of vertices $v \in V$ and $E$ a set of edges that associates each edge $e \in E$ to an unordered pair of vertices $(v_i,v_j)$ for $i=1 \ldots n \wedge j=1 \ldots n$.

3-GCP can be formally defined as a constraint satisfaction problem (CSP) that is represented as the pair $\langle S,\phi \rangle$, where $S=C^n$ with $C^n=\{1,2,3\}$ denotes the free search space, in which all solutions $c \in C^{n}$ are feasible and $\phi$ a Boolean function on $S$ (also a feasibility condition) that divides search space into feasible and unfeasible regions. This function is composed of constraints belonging to edges. In fact, to each $e \in E$ the corresponding constraint $b_e$ is assigned by $b_e(\langle c_1, \ldots , c_n \rangle) = true$ if and only if $e=(v_i,v_j)$ and $c_i \neq c_j$. Assume that $B^i=\{b_e|e=(v_i,v_j) \wedge j=1 \ldots m\}$ defines the set of constraints belonging to variable $v_i$. Then, the feasibility condition $\phi$ is  expressed as a conjunction of all the constraints $\phi (c)= \wedge_{v \in V} B^{v}(c)$.

Typically, constraints are handled indirectly in the sense of the penalty function that transforms the CSP into free optimization problem (FOP)~\cite{Eiben:2003} (also unconstrained problem). Thus, those infeasible solutions that are far away from a feasible region are punished by higher penalties. The penalty function that is also used as a fitness function here, is expressed as:

\begin{equation}
\label{eq:penalty}
 f(c)=min\sum_{i=0}^{n} \psi(c,B^{i}),
\end{equation}

\noindent where the function $\psi(c,B^{i})$ is defined as:

\begin{equation}
\label{eq:viol}
\psi(c,B^{i})=\left\{\begin{matrix}
1 & \textup{if}\ c\ \textup{violates\ at\ least\ one\ } b \in B^{i}, \\
0 & \textup{otherwise}.
\end{matrix}\right.
\end{equation}

In fact, Eq.~(\ref{eq:penalty}) can be used as a feasibility condition in the sense that $\phi(c)=true$ if and only if $f(c)=0$. Note that this equation evaluates the number of constraint violations and determines the quality of solution $c \in C^{n}$.

\section{HABC for Graph 3-coloring}
In the ABC algorithm, the colony of artificial bees consists of three groups~\cite{Yang:2008}: employed bees, onlookers, and scouts. The employed bees discover each food source, that is, only one employed bee exists for each food source. The employed bees share information about food sources with onlooker bees, in their hive. Then, the onlooker bees can choose a food sources to forage. Interestingly, those employed bees whose food source is exhausted either by employed or onlooker bees, becomes scouts. The ABC algorithm is formally described in Algorithm~\ref{alg:prog}, from which it can be seen that each cycle of the ABC search process (statements within a \textbf{while} loop) consists of three functions:

\begin{itemize}
  \item SendEmployedBees(): sending the employed bees onto the food sources and evaluating their nectar amounts,
  \item SendOnlookerBees(): sharing the information about food sources with employed bees, selecting the proper food source and evaluating their nectar amounts,
  \item SendScouts(): determining the scout bees and then sending them onto possibly new food sources.
\end{itemize}

\begin{algorithm}[htb]
\caption{Pseudo code of the ABC algorithm}
\label{alg:prog}
%\scriptsize
%\footnotesize
\begin{algorithmic}[1]
\STATE Init();
\WHILE {!TerminationConditionMeet()}
\STATE SendEmployedBees();
\STATE SendOnlookerBees();
\STATE SendScouts();
\ENDWHILE
\end{algorithmic}
\end{algorithm}

However, before this search process can take place, initialization is performed (function Init()). A termination condition (function TerminationConditionMeet()) is responsible for stoping the search cycle. Typically, the maximum number of function evaluations $MAX\_FES$ is used as the termination condition.

The ABC algorithm belongs to population-based algorithms, where the solution of an optimization problem is represented by a food source. The solution of 3-GCP is represented as a real-valued vector $Y_{i}=\{w_{ij}\}$ for $i=1...\mathit{NP} \wedge j=1...n$, where $w_{ij}$ denotes the weight associated with the $j$-th vertex of the $i$-th solution; $\mathit{NP}$ is the number of solutions within the population, and $n$ the number of vertices. The values of the weights are taken from the interval $w_{ij} \in [lb,ub]$, where $lb$ indicates the lower, and $ub$ the upper bounds. The initial values of the food sources are generated randomly, according to the equation:

\begin{equation}
\label{eq:eq1}
 w_{ij}=\Phi_{ij} \cdot (ub-lb)+lb,
\end{equation}

\noindent where the function $\Phi_{ij}$ denotes the random value from the interval $[-1,1]$.

The employed and onlooker bees change their food positions within the search space, according to the equation:

\begin{equation}
\label{eq:eq2}
 w_{ij}^{'}=w_{ij}+\Phi_{ij} (w_{ij}-w_{kj}),
\end{equation}

\noindent where $\Phi_{ij}$ is a random number from interval $[-1,1]$. The onlooker bee selects a food source with regard to the probability value associated with that food source $p_i$ calculated by the equation:

\begin{equation}
\label{eq:eq3}
p_{i}=\frac{f(\Gamma(Y_{i}))}{\sum_{j=0}^{\mathit{NP}}f(\Gamma(Y_{j}))},
\end{equation}

\noindent where $\Gamma$ indicates a mapping from the real-valued search space to the problem search space, as explained in the next subsection, and $f$ the fitness function according to Eq.~(\ref{eq:penalty}).

\subsection{Fitness calculation}
The ABC for 3-GCP explores continuous real-valued search space, where the solution is represented as $Y_i=\{ w_{ij}\}$ for $i=1...\mathit{NP} \wedge j=1...n$. Firstly, this solution needs to be transformed into a permutation of vertices $X_i=\{v_{ij}\}$. Such a permutation can be decoded into 3-coloring $C_i=\{c_{ij}\}$ by the DSatur heuristic. The 3-coloring $C_{i}$ represents the solution of 3-GCP in its original problem space. Whilst a new position regarding a food source is performed within the real-valued search space, its quality is evaluated within the original problem space, according to the equation~Eq.(\ref{eq:penalty}). This relation can be expressed mathematically as follows:

\begin{equation}
\label{eq:map}
 X_{i}=\Gamma(Y_{i}),\ \ \ \ \ \ \ \   \textup{for}\ i=1...\mathit{NP}.
\end{equation}

Note that the function $\Gamma$ is not injective, i.e. more than one food source can be mapped into the same value of the fitness function. On the other hand, a weakness of this function is that a small move in the real-valued search space can cause a significant increase or decrease in the fitness function.

\subsection{Hybridization with local search}
In the classical ABC algorithm, scouts act as a random selection process. That is, if the position of a food source cannot be improved further within a predetermined number of cycles called $limit$, then that source is replaced by the randomly generated position. In HABC, instead of randomly generating the new position in the search space (exploration), a deterministic exploitation in the vicinity of the current solution was used~\cite{Iacca:2011}. Thus, in place of the original SendScouts() function, the RWDE local search heuristic was implemented, which generates the new food sources according to the following equation~\cite{Rao:2009}:

\begin{equation}
\label{eq:rwde}
 Y_{i}^{'}=Y_{i}+\lambda \cdot U_{i},
\end{equation}

\noindent where $\lambda$ is the prescribed scalar step size length and $U_{i}$ is a unit random vector generated for the $i$-th solution.
%In practice, the ABC for 3-GCP converges very quickly to the promising region of the search space.

\section{Experiments and Results}
The goal of the experimental work was to show that HABC can be successfully applied to 3-GCP. In line with this, the proposed HABC was compared with: EA-SAW, Tabucol, and HEA, whose implementations were downloaded from the Internet.

The characteristics of the HABC used during the experiments were as follows: The population size was set at 100 because this value represents a good selection, as was indicated during the experimental work. The value of $limit$ was set at 1,000, whilst the $MAX\_FES$ was limited to 300,000. The former value was obtained through experimental work, whilst the later was selected in order to draw a fair comparison with the other algorithms, i.e. the other algorithms also obey the same limitation. In the end, 25 independent runs were observed, because of the stochastic nature of observed algorithms.

The algorithms were compared according to two measures: $success\ rate$ (SR) and $average\ number\ of\ objective\ function\ evaluations\ to\ solution$ (AES). The first measure expresses the ratio of the number of successful runs from among all runs, whilst the second reflects the efficiency of a particular algorithm.

\subsection{Test suite}
All graphs in the test suite were generated using the Culberson random graph generator~\cite{Culberson:2011}, which allows to generate graphs of different: size $n$, type $t$, edge probability $p$, and seed $s$. This paper focuses on medium-sized graphs, i.e. graphs with 500 vertices. Three types of graphs were used as follows: uniform (a random graph with variability set at zero), equi-partite, and flat graphs. The edge probability was varied from 0.008 to 0.028 with steps of 0.001. Thus, 21 instances of randomly generated graphs were obtained. Ten different seeds were employed, i.e. from 1 to 10. In summary, each algorithm was solved $3 \times 21 \times 10 \times 25 = 15,750$ different instances of graphs.

An interval of edge probabilities was selected such that the region of $phase$ $transition$ was included. Phase transition is a phenomenon that is connected with most combinatorial optimization problems and indicates those regions, where the problem passes over the state of "solvable" to the state of "unsolvable", and vice versa~\cite{Turner:1988}. The 3-GCP determination of this phenomenon is connected with parameter edge probability. Interestingly, this region is identified differently by many authors. For example, Petford and Welsh~\cite{Petford:1989} stated that this phenomenon occurs when $2pn/3 \approx 16/3$, Cheeseman et al.~\cite{Cheeseman:1991} when $2m/n \approx 5.4$, and Eiben et al.~\cite{Eiben:1998} when $7/n \leq p \leq 8/n$. In our case, the phase transition needed to be by $p=0.016$ over Petford and Welsh, by $p \approx 0.016$ over Cheeseman, and between $0.014 \leq p \leq 0.016$ over Eiben et al..

\subsection{Influence of edge probability}
\begin{figure}[H]	%hbt
\vspace{-5mm}
\centering
\subfigure[SR by uniform graphs] {\includegraphics[width=6.0cm]{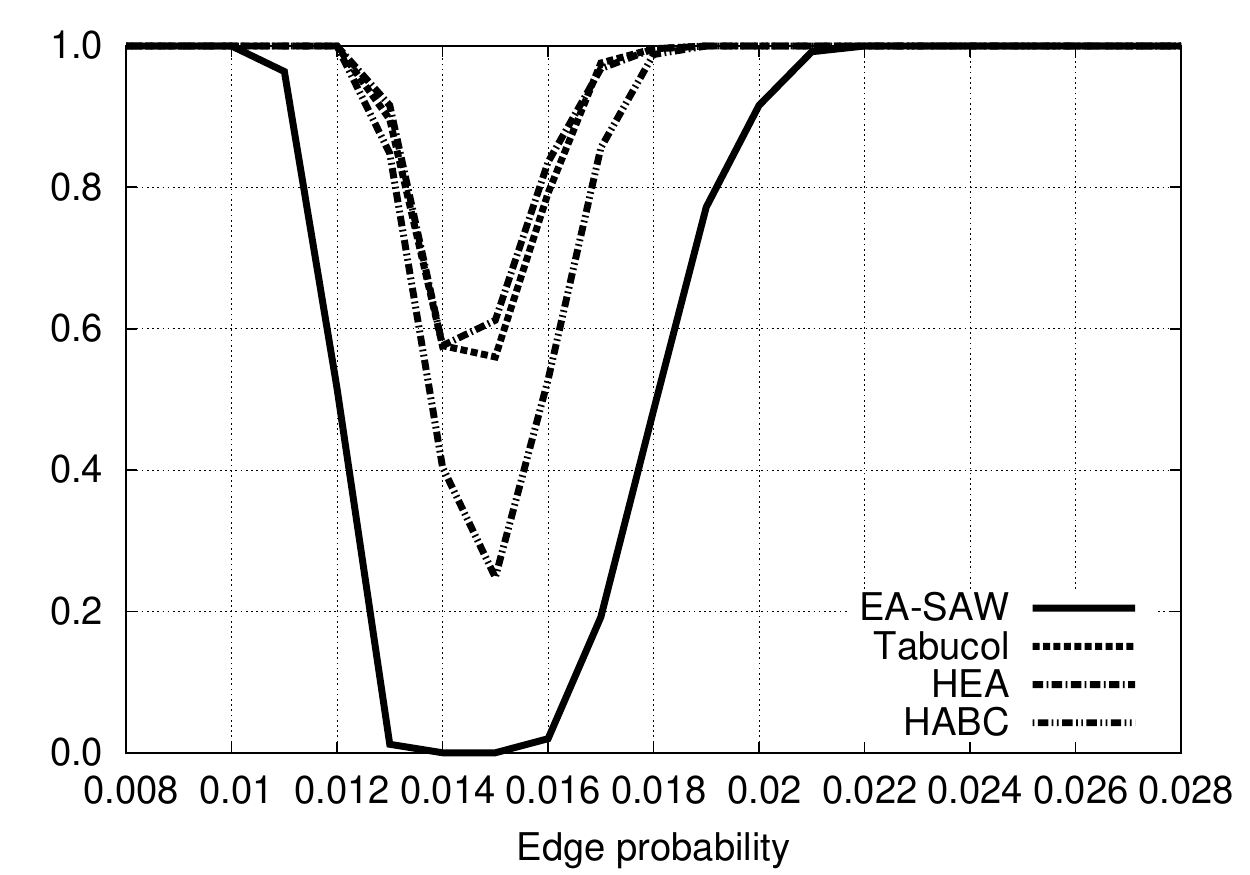}}
\subfigure[AES by uniform graphs] {\includegraphics[width=6.0cm]{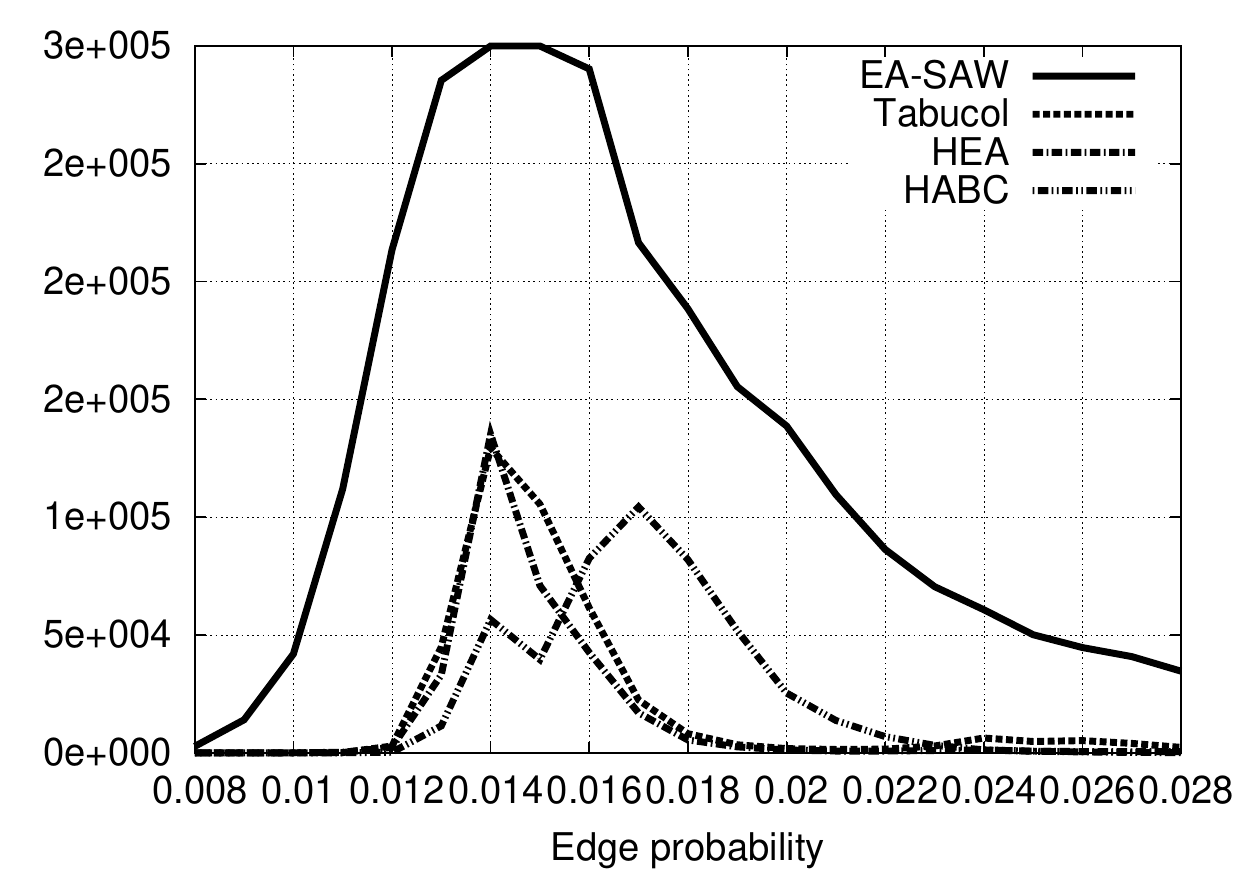}}
\subfigure[SR by equi-partite graphs] {\includegraphics[width=6.0cm]{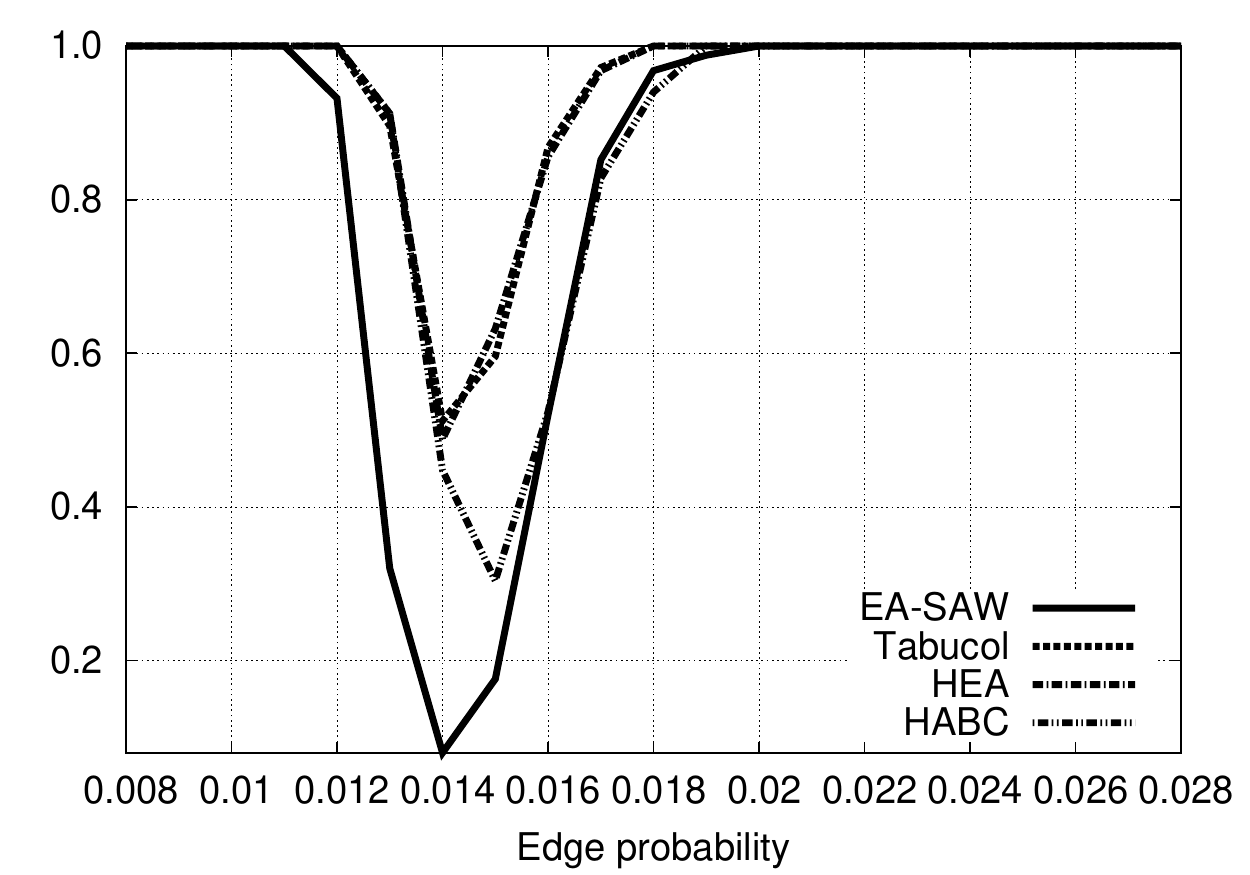}}
\subfigure[AES by equi-partite graphs] {\includegraphics[width=6.0cm]{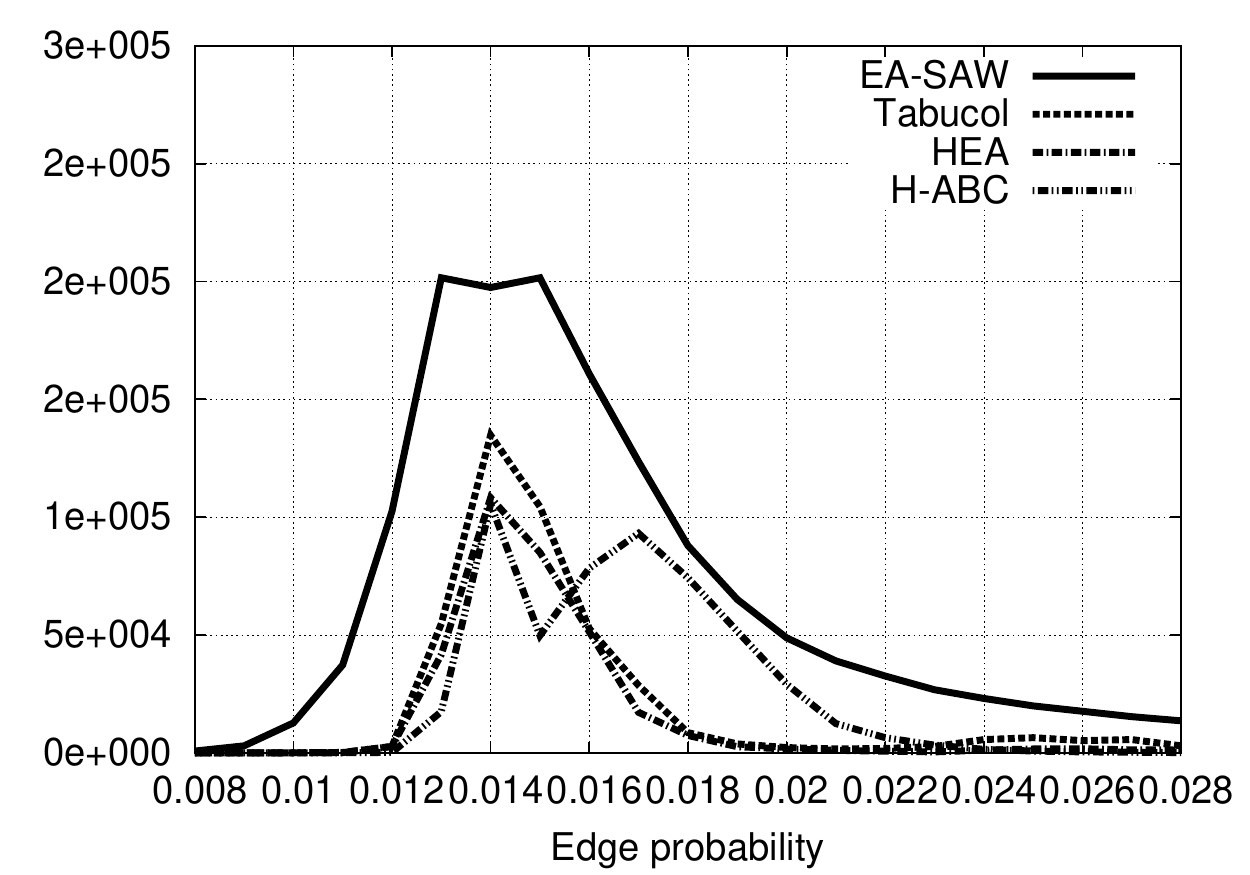}}
\subfigure[SR by flat graphs] {\includegraphics[width=6.0cm]{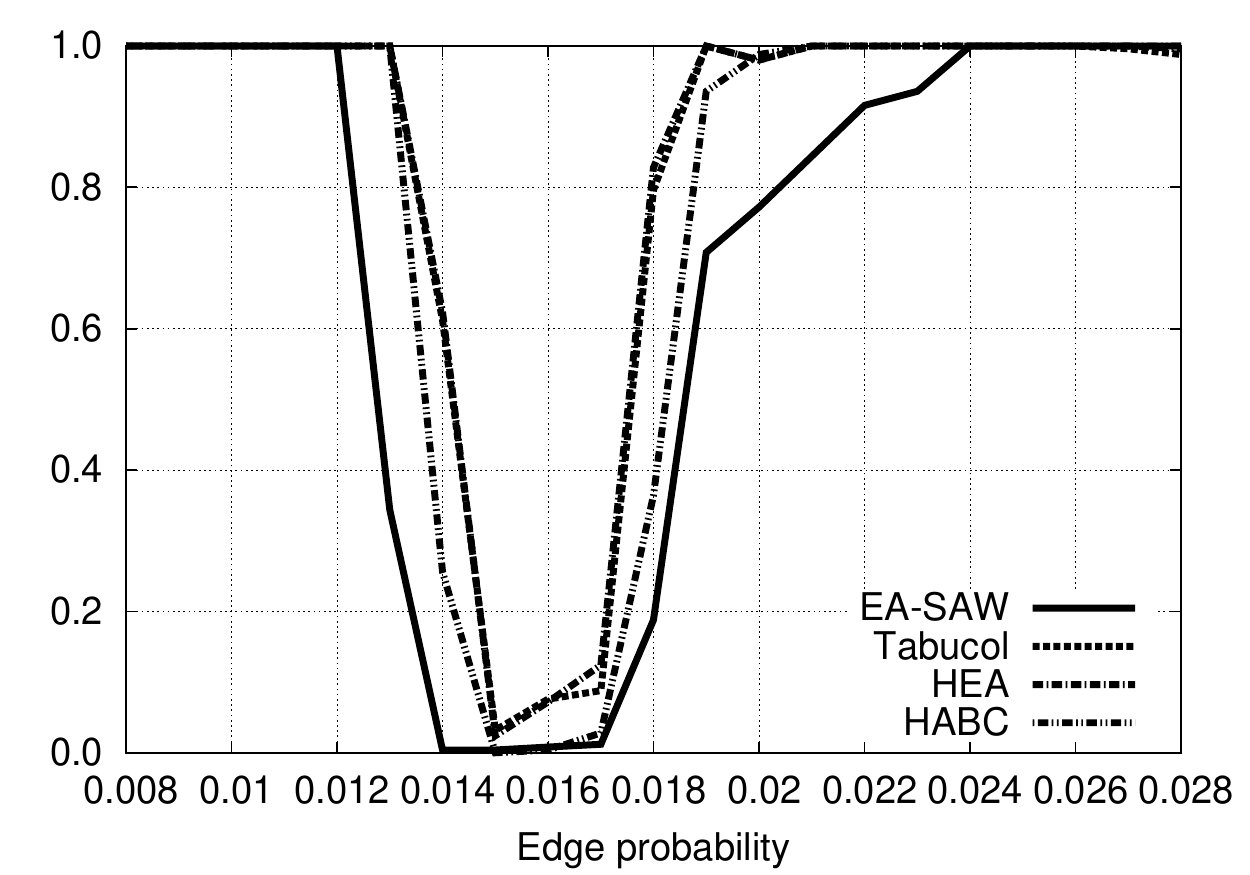}}
\subfigure[AES by flat graphs] {\includegraphics[width=6.0cm]{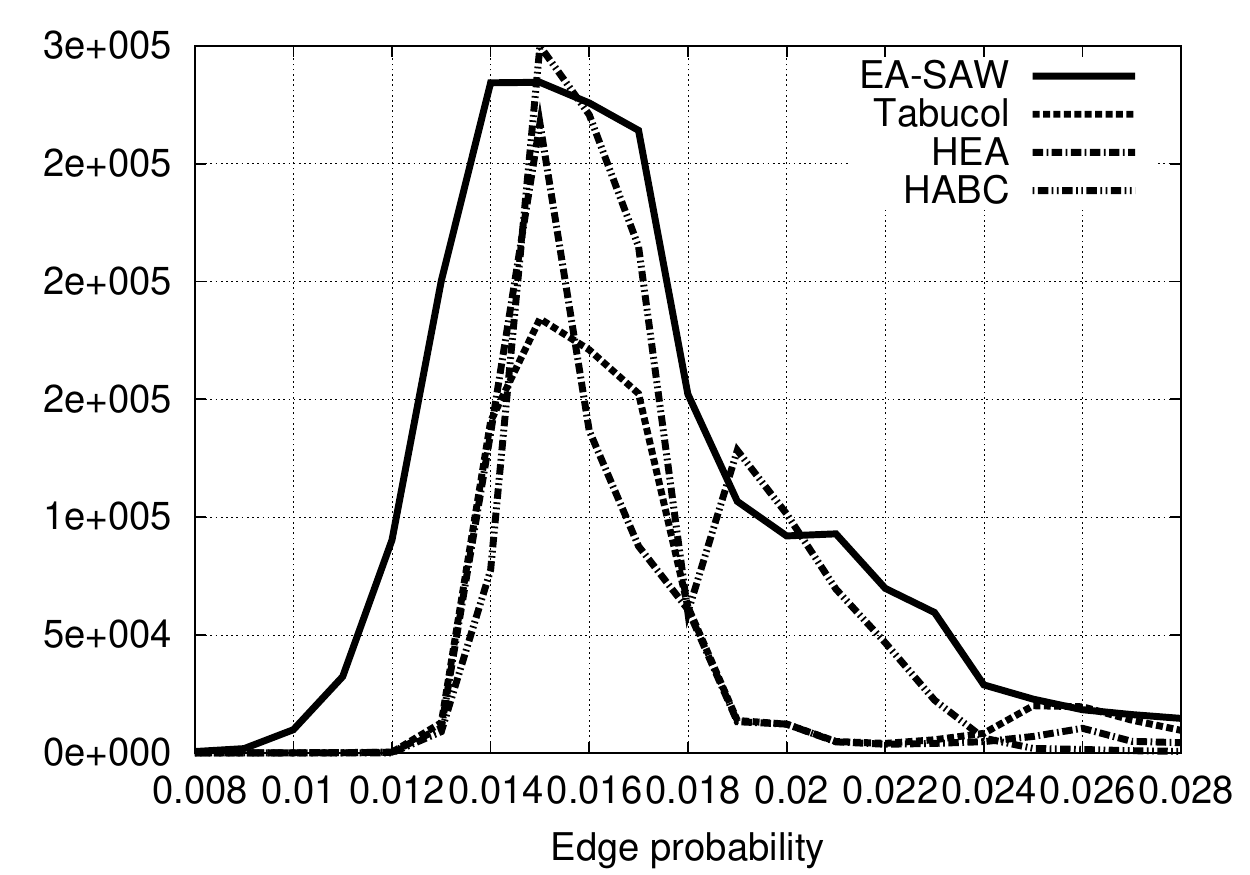}}
\caption{Results of algorithms for 3-GCP solving different types of random graphs}
\label{fig:Sub_1}
\vspace{-5mm}
\end{figure}

In this experiment, the phenomenon of phase transition was investigated, as illustrated by Fig.~\ref{fig:Sub_1}. The figure is divided into six graphs according to type, and two different measures SR and AES. The graphs capture the results of 21 instances that were obtained by varying the edge probability through a region, including phase transition. Due to space limitation of this paper's length, a more detailed analysis of the results is left to the reader.

In summary, the best results on medium-sized graphs were reported by HEA and Tabucol. The results of HABC were slightly worse but comparable to both of the mentioned algorithms, whilst the EA-SAW saw the worst results.

\subsection{Influence of the local search}
During this experiment, the influence of hybridizing the ABC algorithm with a RWDE local search heuristic was observed. Therefore, a especial focus was placed on the instances during phase transition, i.e. $p \in [0.013,0.017]$. The two versions of ABC were compared: the original and the hybridized version. In the former, the scouts were generated randomly, whilst in the later the RWDE local search heuristic was used.

The results of this experiment are shown in Table~\ref{tab:tab_1}, where the row $Graphs$ indicates different graph types, whilst the columns $Random$ and $Hybrid$ indicate the original and the hybrid ABC algorithms. Note that the average results of the mentioned instances have a varying seed $s \in [1,10]$ and are presented in the table.

\begin{table}[htb]        %[!htb]
\caption{Influence of the local search by HABC on different graph types.}
\label{tab:tab_1}
\vspace{-5mm}
\small
\begin{center}
\newcolumntype{R}{>{\raggedleft\arraybackslash}X}
\begin{tabularx}{\textwidth}{ c  R  R  R  R  R  R }
\hline
 Graphs & \multicolumn {2}{c}{Uniform} & \multicolumn {2}{c}{Equi-partite} & \multicolumn {2}{c}{Flat}  \\ \cline{ 2-7}
 p & Random & Hybrid & Random & Hybrid & Random  & Hybrid  \\
\hline
0.013 & 0.816 & 0.848 & 0.872 & 0.912 & 1.000 & 1.000 \\
0.014 & 0.112 & 0.404 & 0.200 & 0.448 & 0.012 & 0.256 \\
0.015 & 0.060 & 0.248 & 0.036 & 0.304 & 0.000 & 0.000 \\
0.016 & 0.180 & 0.528 & 0.104 & 0.524 & 0.000 & 0.004 \\
0.017 & 0.328 & 0.856 & 0.340 & 0.828 & 0.000 & 0.028 \\
\hline
avg   & 0.299 & 0.577 & 0.310 & 0.603 & 0.202 & 0.258 \\
\hline
\end{tabularx}
\end{center}
\vspace{-10mm}
\normalsize
\end{table}

The results showed that using the RWDE local search, substantially improved the results of the original ABC. For example, this improvement amounted to $92.98\%$ for uniform, $94.52\%$ for equi-partite, and $27.72\%$ for flat graphs. On average, hybridization improved the results of the original ABC for $71.74\%$.

\section{Conclusion}
The results of the proposed HABC for 3-GCP convinced us that the original ABC algorithm is a powerful tool for solving combinatorial optimization problems. HABC gained results that are comparable with the results of the best algorithm for $k$-GCP today (Tabucol and HEA), and improved results obtained with EA-SAW when solving the medium-sized extensive suite of random generated graphs. Note that these graphs are not the hardest to color but are difficult enough that the suitability of the ABC technology for solving the 3-GCP could be successfully proven.

In the future, the HABC for 3-GCP could be additionally improved. In particular, the problem-specific knowledge via local search heuristics could be conducted into the algorithm. The greatest challenge for further work remains the solving of large-scale graph suite (graphs with 1,000 vertices). We are convinced that these graphs could also be successfully solved using the proposed HABC algorithm.

\bigskip{\small \smallskip\noindent Updated 5 May 2012.}
\end{document}